\documentclass[sigconf]{acmart}

\usepackage{balance}
\AtBeginDocument{%
  }

\copyrightyear{2023}
\acmYear{2023}
\setcopyright{acmlicensed}\acmConference[MM '23]{Proceedings of the 31st ACM International Conference on Multimedia}{October 29-November 3, 2023}{Ottawa, ON, Canada}
\acmBooktitle{Proceedings of the 31st ACM International Conference on Multimedia (MM '23), October 29-November 3, 2023, Ottawa, ON, Canada}
\acmPrice{15.00}
\acmDOI{10.1145/3581783.3612489}
\acmISBN{979-8-4007-0108-5/23/10}

\begin{document}

\title{Control3D: Towards Controllable Text-to-3D Generation}


\author{Yang Chen}
\affiliation{%
  \institution{University of Science and Technology of China}
  \country{China}
  }
\email{c1enyang.ustc@gmail.com}

\author{Yingwei Pan}
\affiliation{%
  \institution{University of Science and Technology of China}
    \country{China}
  }
\email{panyw.ustc@gmail.com}

\author{Yehao Li}
\affiliation{%
  \institution{Sun Yat-sen University}
    \country{China}
  }
\email{yehaoli.sysu@gmail.com}

\author{Ting Yao}
\affiliation{%
  \institution{HiDream.ai Inc.}
    \country{China}
  }
\email{tingyao.ustc@gmail.com}

\author{Tao Mei}
\affiliation{%
  \institution{HiDream.ai Inc.}
    \country{China}
  }
\email{tmei@hidream.ai}



\renewcommand{\shortauthors}{Yang Chen, Yingwei Pan, Yehao Li, Ting Yao, \& Tao Mei}

\begin{abstract}
Recent remarkable advances in large-scale text-to-image diffusion models have inspired a significant breakthrough in text-to-3D generation, pursuing 3D content creation solely from a given text prompt. However, existing text-to-3D techniques lack a crucial ability in the creative process: interactively control and shape the synthetic 3D contents according to users’ desired specifications (e.g., sketch). To alleviate this issue, we present the first attempt for text-to-3D generation conditioning on the additional hand-drawn sketch, namely \textit{Control3D}, which enhances controllability for users. In particular, a 2D conditioned diffusion model (ControlNet) is remoulded to guide the learning of 3D scene parameterized as NeRF, encouraging each view of 3D scene aligned with the given text prompt and hand-drawn sketch. Moreover, we exploit a pre-trained differentiable photo-to-sketch model to directly estimate the sketch of the rendered image over synthetic 3D scene. Such estimated sketch along with each sampled view is further enforced to be geometrically consistent with the given sketch, pursuing better controllable text-to-3D generation. Through extensive experiments, we demonstrate that our proposal can generate accurate and faithful 3D scenes that align closely with the input text prompts and sketches.
\end{abstract}


\begin{teaserfigure}
\vspace{-0.18in}
\centering
  \includegraphics[width=0.93\textwidth]{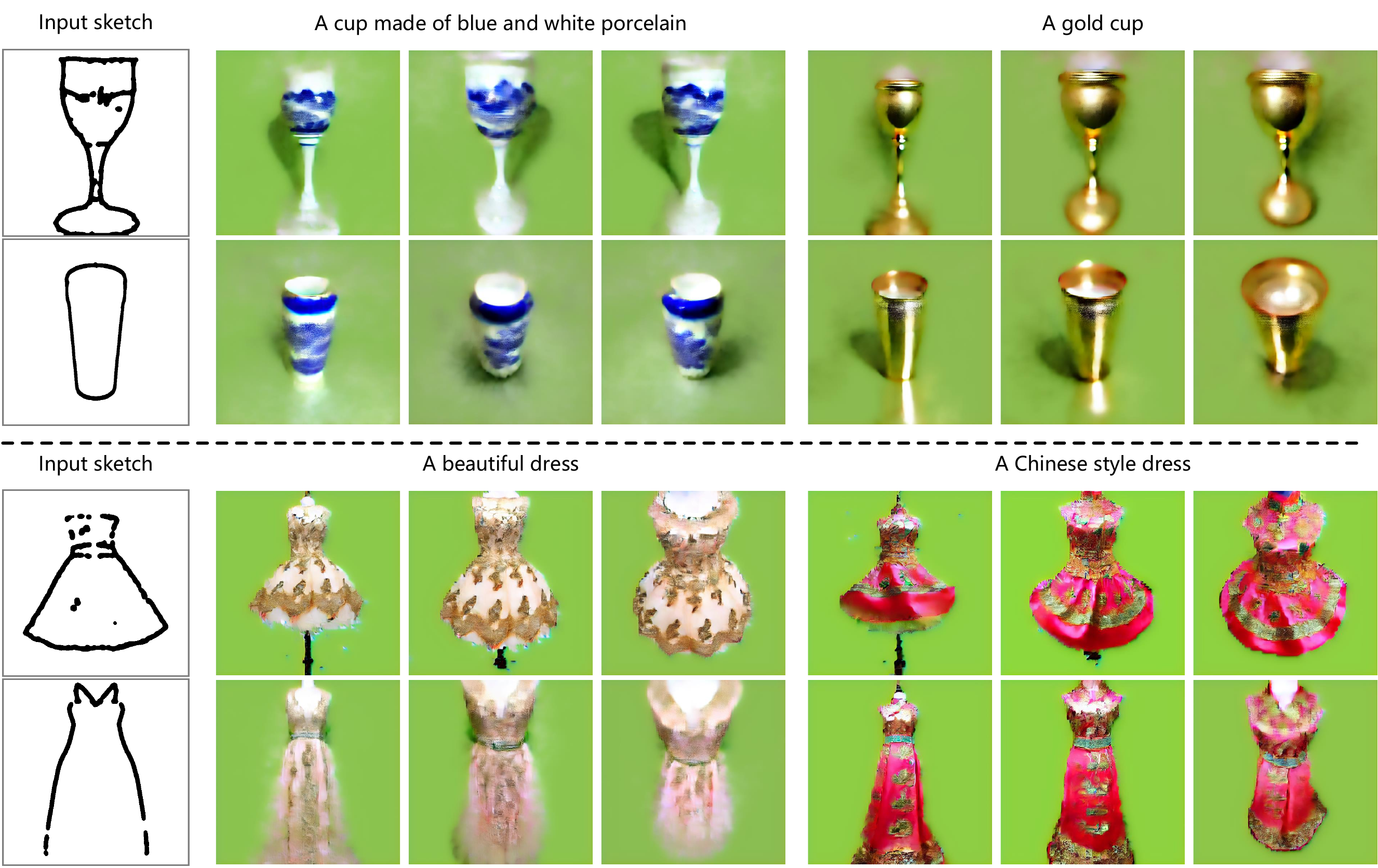}
    \vspace{-0.1in}
    \caption{We propose \textit{Control3D}, a new model that can control text-to-3d generation with an additional hand-drawn sketch. }
  \vspace{-0.0in}
  \label{fig:teaser}
\end{teaserfigure}


\maketitle

\section{Introduction}
3D content creation has significantly impacted the multimedia field by enabling the construction of immersive and engaging digital worlds. The applications from video games and animated films to virtual reality present vast massive opportunities for 3D professionals to deliver compelling experiences that captivate audiences. The conventional process of 3D content creation is commonly time-consuming, necessitating the expertise of professional designers with significant experience in 3D software tools. As such, there is a strong urge to automate this creative process, not only for enhancing the efficiency of 3D designers with domain expertise but also for democratizing 3D content creation for novices.

Recent advancements in 3D-specific generative adversarial networks (GANs) \cite{nguyen2019hologan, chan2022efficient, liao2020towards, niemeyer2021giraffe, chan2021pi, gu2021stylenerf} have reduced the technical barriers to creating 3D content. However, constructing such GAN models requires cost-expensive large-scale 3D data collection and meticulous pre-processing. Therefore, these models are typically confined to a pre-defined single object category, which severely restricts the diversity of synthetic 3D content and their practical applicability. In contrast, more recent text-to-3D generation works \cite{poole2022dreamfusion, lin2022magic3d, wang2022score, metzer2022latent} demonstrated their remarkable ability to create 3D scenes solely based on human-written text prompts, yielding various impressive 3D assets. Such automatic 3D content creation from input text prompts alone can greatly emancipate ordinary users from the need to acquire all the skills required for creating 3D assets. Nevertheless, simplifying the 3D generation interface as a text-only format may impede users' ability to fully articulate their desired specifications (e.g., visual prompts like sketch). Accordingly, it is crucial to explore more robust interfaces that offer comprehensive control signals for 3D content creation.

In this work, we propose \textit{Control3D}, the first attempt that enhances users' controllability by upgrading text-to-3D generation with additional hand-drawn sketch conditions. The incorporation of sketching aligns with human's innate ability in drawing and painting, offering a more intuitive and natural way for users to interactively control 3D content creation. To achieve this, we draw inspiration from recent works \cite{poole2022dreamfusion, wang2022score} which leverage large-scale pre-trained text-to-image diffusion models to optimize a Neural Radiance Field (NeRF) by applying score distillation sampling (SDS). SDS estimates the optimization direction of NeRF such that the distributions of rendered images derived from the 3D model are pushed to a higher density probability region determined by the input text prompt. Consequently, the generated 3D scenes are semantically aligned with the input text prompts. Herein, we take one step further by extending the typical text-driven SDS with more conditions of sketch. In particular, we propose to integrate the optimization of NeRF with an image-conditioned text-to-image diffusion model (ControlNet \cite{zhang2023adding}), which triggers the control of diffusion models with additional sketch conditions. Although this way simply guides text-to-3D generation with the control signals of sketches, we observe that such implicit control process is insufficient to produce high-quality 3D scene that precisely maintains the same geometric structure of given sketch.

To alleviate this issue, we further design a novel sketch consistency loss that explicitly encourages the geometric consistency between synthetic 3D scene and given sketch. Technically, in each training step, we utilize a pre-trained differentiable photo-to-sketch model \cite{li2019photo} to estimate the sketch of the rendered image. We then employ the sketch consistency loss to match the embeddings between the estimated sketch and the given input sketch. This design encourages NeRF model to generate outputs that adhere to the sketch-specified geometry from arbitrary poses.

We conduct thorough experiments to verify the effectiveness of our proposed method. Experimental results show that our proposed \textit{Control3D} is capable of generating realistic 3D scenes with remarkable likeness to the given sketch while also respecting the contexts present in the input text prompt. 
In sum, we have made the following contributions:
\begin{itemize}
    \item We propose \textit{Control3D}, a new framework to create realistic 3D scene conditioned on a text prompt and a visual prompt (sketch image). To the best of our knowledge, this is the first attempt to control text-to-3D generation with a human-drawn sketch. 
    \item We additionally introduce a novel sketch consistency loss that explicitly enforces the synthetic 3D scene to precisely preserve the same geometric structure as in the given sketch.
    \item We perform extensive experiments to demonstrate that our controllable text-to-3D generation results not only have plausible appearances and shapes, but also faithfully conform to the given prompt and sketch.
\end{itemize}

\section{Related Work}
\textbf{Diffusion models.} Diffusion models \cite{sohl2015deep, ho2020denoising, nichol2021improved, ho2022classifier} have emerged as the new trend of generative models for generating diverse, high-quality content. Especially, they have recently been used to form state-of-the-art text-to-image (T2I) models (such as DALL-E2 \cite{ramesh2022hierarchical} and Imagen \cite{saharia2022photorealistic}) with the help of large-scale datasets. These models can generate high-quality images of objects and scenes that are aligned with a natural language text prompt given by the user. In order to reduce the computation resources and improve the inference speed, Latent Diffusion Model (LDM) was further proposed \cite{rombach2022high}. Motivated by the success of these models, many works attempt to control pre-trained T2I diffusion models to support additional input conditions. Textual Inversion \cite{gal2022image} and DreamBooth \cite{ruiz2022dreambooth} are proposed to personalize the contents in the generated images using a small set of images with the same subjects. Recent work ControlNet \cite{zhang2023adding} proposes to control large image diffusion model (i.e., Stable Diffusion) by additional condition inputs like edge maps, segmentation maps, sketches, \textit{etc}. Despite the advances in controllable 2D T2I generation, using text prompts and additional condition images to describe and control 3D generation remains an open and challenging problem in the multimedia field.

\textbf{Text-to-3D generation.}
Recently, significant advancements have been made in multimedia content generation \cite{ho2020denoising, ramesh2022hierarchical, saharia2022photorealistic, chen2019mocycle, chen2019animating, poole2022dreamfusion, jain2022zero, pan2017create, zhang2023}. In between, with recent notable advancements in T2I generation and NeRF based 3D reconstruction, there has been a growing interest in text-to-3D generation. While large-scale paired text-image data is available for T2I generation, paired text-3D data is currently unavailable on a similar scale. To liberate the need for training data, DreamField \cite{jain2022zero} and CLIP-Mesh \cite{mohammad2022clip} leverage cross-modal knowledge from a pre-trained image-text model (i.e., CLIP model) to optimize underlying 3D representations (NeRFs and Meshes). However, these models tend to produce less photorealistic 3D results. More recently, sparked by the success of diffusion models in 2D image generation, DreamFusion \cite{poole2022dreamfusion} and SJC \cite{wang2022score} utilize pre-trained T2I diffusion models for text-to-3D generation and demonstrates impressive results. Following work Magic3D \cite{lin2022magic3d} further improves the generation quality with a coarse-to-fine strategy that leverages both low- and high-resolution diffusion priors. 

Existing methods \cite{jain2022zero, poole2022dreamfusion} can generate 3D assets matching the input text prompt. However, they are unable to control the 3D generation process with additional freehand interfaces. In this paper, we instead tackle a novel and challenging problem, which is to control the text-to-3D generation process with a hand-drawn sketch. Latent-NeRF \cite{metzer2022latent} is perhaps the most related work that uses a 3D mesh as an additional constraint to guide the generation process. However, the 3D mesh is too complex and difficult for ordinary users to design and produce. In contrast, our proposed control signal, presented in the form of a sketch, is more intuitive and user-friendly, making it a more accessible method for users to interact with and control the 3D generation.

\textbf{Sketch-based visual synthesis.} Using a sketching interface to guide computers to generate content can be traced back to Ivan Sutherland's SketchPad \cite{sutherland1964sketch}. This tradition has continued in the area of sketch-based visual synthesis. One common approach is using GANs to learn the mapping between sketches and images. SketchyGAN \cite{chen2018sketchygan} presents an edge-preserving data augmentation technique to train a GAN that can synthesize plausible images from sketches. ContextualGAN \cite{lu2018image} proposes to learn the joint distribution of sketch and image for faithful sketch-to-image generation. Recent works \cite{voynov2022sketch, zhang2023adding} involve pre-trained T2I models in sketch-based visual synthesis. Given a sketch and a text prompt, these models use the sketch to control the diffusion model, producing results that align with the text prompt and follow the spatial layout of the sketch. In this work, we go one step further and make the first attempt that demonstrates sketch controlling in the realm of text-to-3D generation. We notice a related work Sketch2Mesh \cite{guillard2021sketch2mesh} that focuses on 3D generation from sketches. However, our work targets creating realistic 3D scenes conditioned on a text prompt plus a visual prompt (sketch image), which is more challenging.


\textbf{NeRF with Regularizations.}
Recently, Neural Radiance Fields (NeRF) \cite{mildenhall2020nerf} has received significant attention due to its powerful representation ability for 3D scenes. Although NeRF achieves state-of-the-art performance in view synthesis, its ability to reconstruct scenes from a sparse set of input views is significantly limited. The performance drops severely when only a few input views are available. Various external regularizations have been proposed to address this problem \cite{deng2022depth, jain2021putting, kim2022infonerf, niemeyer2022regnerf, xu2022sinnerf}. 
Specifically, DietNeRF \cite{jain2021putting}  introduces semantic consistency constraints that align input and novel views. InfoNeRF \cite{kim2022infonerf} proposes a ray entropy minimization regularization to implicitly regularize the density field, while DS-NeRF \cite{deng2022depth} explicitly incorporates additional depth supervision. 
RegNeRF \cite{niemeyer2022regnerf} introduces a normalizing flow $\&$ depth smoothness regularization, and SinNeRF \cite{xu2022sinnerf} proposes multiple semantic and geometry regularizations in a semi-supervised perspective. These advancements are highly significant in the development of NeRF and provide valuable insights for the field of view synthesis. However, the aforementioned regularizations often rely on ground truth views of the 3D scene, while our \textit{Control3D} focuses on a more challenging setting that only has a text prompt and sketch as input. In order to pursue better controllable text-to-
3D generation, we design a novel sketch consistency loss to regularize the NeRF optimization.

\begin{figure*}
    \centering
    \includegraphics[width=1.\linewidth]{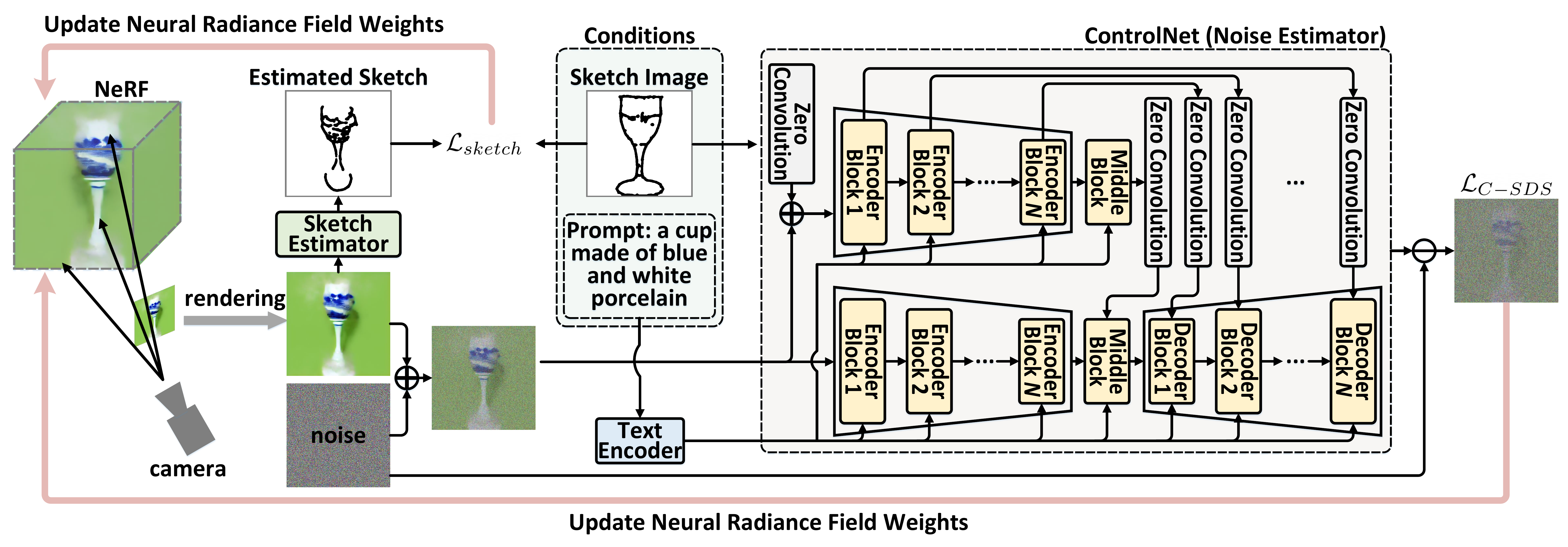}
    \vspace{-0.28in}
    \caption{\textbf{An overview} of \textit{Control3D}. Given a text prompt and a hand-drawn sketch as input, our method can generate plausible 3D content that faithfully matches the input conditions. Similar to Dreamfusion \cite{poole2022dreamfusion} and SJC \cite{wang2022score}, we first use Neural Radiance Field (NeRF) to represent the 3D scene. In each iteration, we render an image from a random viewpoint. The image is perturbed into a noisy sample to input into ControlNet \cite{zhang2023adding}. The input sketch image and text prompt are also input into ControlNet as conditions. Then conditioned score sampling distillation loss $\mathcal{L}_{C-SDS}$ is calculated and the NeRF model is updated with it to closely align with the text prompt and sketch. Additionally, we utilize a photo-to-sketch model (sketch estimator) \cite{li2019photo} to estimate the sketch of the rendered image. Then we compute $\mathcal{L}_{sketch}$ to ensure NeRF maximally maintains the geometry cues of the input sketch by constraining the estimated sketch's CLIP \cite{radford2021learning} embedding to be similar to the input sketch's. 
    }
    \vspace{-0.2in}
    \label{fig:framework}
\end{figure*}

\section{Method}
In this section, we elaborate our proposed \textit{Control3D}, which leverages a hand-drawn sketch to guide text-to-3D generation. Our approach provides an intuitive and user-friendly control for text-to-3D generation. We start by briefly reviewing the background of Neural Radiance Fields and diffusion models. We then continue to introduce our method and how we apply the image-conditioned 2D diffusion model and novel sketch consistency loss functions to enable controllable text-to-3D generation. Figure \ref{fig:framework} depicts an overview of our \textit{Control3D} model.

\subsection{Background}
\textbf{Neural Radiance Fields.} Neural Radiance Fields (NeRF) \cite{mildenhall2020nerf} provides a compact and convenient representation for 3D scenes, which achieves impressive results on novel view synthesis. A typical NeRF can be parameterized as a function $F_{\theta} : (\mathbf{p}, \mathbf{d}) \to (\mathbf{c},\sigma)$, which maps a 3D location $\mathbf{p} \in \mathbb{R}^3$ and viewing direction $\mathbf{d} \in \mathbb{S}^2$ into a volume density $\sigma \in [0, \infty)$ and color value $\mathbf{c} \in [0, 1]^3$. To compute the color of a single pixel, NeRF integrates color along rays cast from the observer according to volume rendering \cite{max1995optical}:
\begin{equation}
\label{eq:nerf_c}
\hat{C}(\mathbf{r}) = \int^{k_f}_{k_n}W(k) \sigma(\mathbf{r}(k))\mathbf{c}(\mathbf{r}(k), \mathbf{d}) \mathrm{d}k,
\end{equation}
where the ray $\mathbf{r}(k) = \mathbf{o} + k\mathbf{d}$ originating at the camera center $\mathbf{o}$ through the pixel along direction $\mathbf{d}$, and the accumulated transmittance $W(k)=\exp(-\int^{k}_{k_n}\sigma(\mathbf{r}(s)ds)$ weights the radiance by the ray travels from the image plane at $k_n$ to $k$ unobstructed. To approximate the integral, NeRF employs a hierarchical sampling algorithm to select points within near and far bound $k_n$ and $k_f$ along each ray. Since all processes are fully differentiable, NeRF training loss is formulated as a pixel-wise photometric reconstruction error between rendered pixel color and the ground truth color $\mathcal{L}(\hat{C}(\mathbf{r}), C(\mathbf{r}))$. To render an image, a collection of rays are sampled corresponding to all the pixels in that image, and the resulting color values $\hat{C}(\mathbf{r})$ are arranged into a 2D image.

\textbf{Diffusion Models.} Diffusion models (DMs) are generative models that can generate samples from a Gaussian distribution to match target data distribution by a gradual denoising process \cite{ho2020denoising}. 
In the forward diffusion process $q(\cdot)$, Diffusion models gradually add Gaussian noises to a ground truth image $x_0$ according to a predetermined schedule $\beta_1,\beta_2,...,\beta_T$: 
\begin{equation}
    q(x_t|x_{t-1})= \mathcal{N}(x_t;\sqrt{1-\beta_t}x_{t-1}, \beta_t\bf{I}),
\end{equation}
where $x_t$ is a noised ample with noise level $t$. The reverse process consists of denoising steps that progressively remove noise by modeling a neural network $\epsilon_\phi$ with parameters $\phi$ that predicts the noise $\epsilon$ contained in a noisy image $x_t$ at step $t$. The loss function for training the diffusion model is formulated as follows:
\begin{equation}
\label{eq:diff}
\mathcal{L}_{diff}(\phi, x) = \mathbb{E}_{t, \epsilon} [w(t)\parallel\epsilon_\phi(x_t,t) - \epsilon \parallel_2^2],
\end{equation}
where t uniformly sampled from $\{1,...,T\}$ and $w(t)$ is a weighting function that depends on the timestep $t$. Then $x_{t-1}$ can be reconstructed from $x_t$ by removing the predicted noise:
\begin{equation}
\label{eq:backward}
x_{t-1} = \frac{1}{\sqrt{\alpha_t}} (x_t - \frac{1 - \alpha_t}{\sqrt{1 - \bar{\alpha}_t}} \epsilon_\phi(x_t, t)) + \eta_t \epsilon,
\end{equation}
where $\alpha_t = 1 - \beta_t$, $\bar{\alpha}_t = \prod \limits_{s=1}^t \alpha_s$ and $\eta_t^2 = \beta_t$ , $\epsilon \sim \mathcal{N}({\bf{0,I}})$.

A T2I diffusion model builds upon the above theory and receives a text prompt as an additional condition. Given a text prompt $y$, a text encoder first maps it into text embedding. Then the text embedding is injected into the diffusion model via attention mechanism widely adopted in Vision Transformers \cite{yao2023dual,li2022contextual,yao2022wave}. Formally, the T2I diffusion model can be denoted as $\epsilon_{\phi}(x_t,t,y)$.

\subsection{Control3D}
\label{sec:control3d}
In pursuit of facilitating controllable text-to-3D generation, our method takes hand-drawn sketches as an additional condition to guide text-to-3D generation. In this section, we first describe how we integrate text-to-3D generation with a 2D conditioned diffusion model (ControlNet) in the process of text-to-3D generation, then describe our proposed sketch consistency loss for pursuing better controllable text-to-3D generation.

\textbf{Text-to-3D Generation with Score distillation Sampling.} A recent pioneering practice (Dreamfusion \cite{poole2022dreamfusion}) designs Score distillation sampling (SDS), which enables utilizing a pre-trained T2I diffusion
model to optimize a NeRF model solely based on a text prompt $y$. Formally, let the NeRF model parameterized by $\theta$ and $g_{\theta}(\pi)$ be a differentiable volumetric renderer that can produce an image $x$ at a given camera pose $\pi$, \textit{i.e.,} $x=g_{\theta}(\pi)$. The SDS loss provides the gradient direction to update NeRF parameters $\theta$:
\begin{equation}
\label{eq:sds}
\nabla_{\theta}\mathcal{L}_{SDS}(\phi, x) = \mathbb{E}_{t,\epsilon}[ w(t)( \epsilon_\phi(x_t;t,y)-\epsilon)\frac{\partial x}{\partial \theta}]. 
\end{equation}
The SDS loss perturbs the rendered image $x$ (\textit{i.e.,} one view of the NeRF's output) into a noisy sample $x_t$ at arbitrary timestep $t$ as described in the forward diffusion process. Then, $x_t$ and the input text $y$ are taken as inputs of diffusion model to predict the noise $\epsilon_\phi(z_t,t,y)$, which should be the same as the added noise $\epsilon$. Intuitively, by doing so, SDS loss pushes the rendered images towards the higher-density regions under the text-conditioned diffusion prior, \textit{i.e.,} to be realistic and resemble the given input text prompt.

\textbf{Sketch-controlled Text-to-3D Generation.} Given a sketch image $I_s$ and a text prompt $y$, our goal is to generate a realistic 3D scene that not only follows the sketch outline but also respects the contexts present in the input text prompt. To achieve this goal, we remould the standard SDS based text-to-3D pipeline by exploiting an image conditioned diffusion model (ControlNet \cite{zhang2023adding}) to trigger sketch-controlled text-to-3D. ControlNet is an end-to-end neural network architecture that controls a large-scale pre-trained image diffusion model (Stable Diffusion) to learn task-specific input conditions. Specifically, herein we use ControlNet-scribble\footnote{https://huggingface.co/lllyasviel/sd-controlnet-scribble} as our diffusion prior model, which is trained on large-scale sketch-image-text pairs and can enable sketch-guided text-to-image generation. 
As ControlNet-scribble has two conditions, sketch $I_s$ and text prompt $y$, the noise is estimated as follows:
\begin{equation}
\begin{aligned}
\hat{\epsilon}_\phi(x_t; t, y, I_s) = &\epsilon_\phi(x_t; t, y, \lambda*I_s) \\ &+ s * (\epsilon_\phi(x_t; t, y, \lambda*I_s) - \epsilon_\phi(x_t; t)),
\end{aligned}
\end{equation}
where $s$ is the scale of classifier-free guidance \cite{ho2022classifier} and $\lambda \in [0, 1]$ is a hyper-parameter that determines the control degree of the conditioned sketch image $I_s$. Note that when $\lambda = 0$, the ControlNet-scribble is degraded as a Stable Diffusion model, which will generate images only from the text prompt while ignoring the sketch condition. In this way, our proposed method elegantly incorporates the input sketch image and text prompt in a unified fashion. Similar to Eq. \ref{eq:sds}, we update the NeRF model by the following gradient:
\begin{equation}
\label{eq:c-sds}
\nabla_{\theta}\mathcal{L}_{C-SDS}(\phi, x) = \mathbb{E}_{t,\epsilon}[ w(t)(\hat{\epsilon}_\phi(x_t;t,y,I_s)-\epsilon)\frac{\partial x}{\partial \theta}],
\end{equation}
where $\phi$ is the parameters of the pre-trained ControlNet-scribble. 

Intuitively, previous SDS-based text-to-3D generation (Eq. \ref{eq:sds}) ensures rendered views of the 3D scene lie in a higher probability density region conditioned on a single text prompt under the diffusion prior. However, our conditioned SDS (Eq. \ref{eq:c-sds}) encourages the rendered images also align with the input sketch, where the probability density region is further narrowed down by the sketch condition. As a result, we can obtain a 3D scene that aligns closely with the input text prompt and sketch. Following \cite{poole2022dreamfusion, wang2022score}, our diffusion loss also employed with view-dependent prompting (e.g., adding ``front view'', ``side view'', or ``back view'' with respect to the camera position to the main prompt). We set $\lambda = 1$ for the sketch image corresponded view and $\lambda=0$ for other views to avoid the learned 3D scene being overfitted to the viewpoint of the input sketch image. Nevertheless, through our experiments, we found that implicitly controlling 3D generation solely using 2D diffusion prior commonly fails to ensure the generated 3D scene precisely aligns with the geometry cues of the input sketch.

\textbf{Sketch Consistency Loss.} To mitigate the aforementioned issues, we propose a novel sketch consistency loss to encourage the geometry described by the input sketch to be highly preserved in the 3D generation. 
One intuitive way to achieve this goal is to leverage the input sketch image to directly constrain NeRF rendered images. Nevertheless, the target NeRF rendered images are photo-realistic and thus have a huge domain/style gap with the input sketch image. Thus, it is not trivial to directly encourage the similarity between rendered images and input sketch image. 
In contrast, we propose to utilize an off-the-shelf photo-to-sketch model $G$ \cite{li2019photo} to estimate the sketch of the rendered images. By doing so, we can compare the estimated sketch with the input sketch, thereby easily encouraging the synthetic results geometrically consistent with input sketches. 

Next, a natural solution to compare the ground-truth input sketch with the estimated sketch is to use traditional mean squared error loss. However, such pixel-wise comparison might be misleading. This is because the comparison is only accurate when the estimated sketch is perfectly aligned with the original sketch image's pose, which is often not accessible. Instead, an alternative way is to generally compare the semantic-level representation of the input sketch and estimated sketches captured from different viewpoints. To fulfill this goal, inspired by \cite{jain2021putting}, we utilize a CLIP image encoder $E$ \cite{radford2021learning} to extract normalized image embeddings of the estimated sketch and input sketch. On the one hand, the CLIP image encoder is trained on hundreds of millions of web images that allow the network to understand sketch modality images \cite{radford2021learning, sain2023clip}. On the other hand, it can capture consistent semantic-level representation across varied viewpoints \cite{jain2021putting}.
Then our sketch consistency loss can be formulated by minimizing their cosine similarity:
\begin{equation}
\label{eq:sketch}
\mathcal{L}_{sketch}=-E(G(x))^{T}E(I_s),
\end{equation}
where $x$ is a rendered image by the NeRF model from an arbitrary viewpoint. Although there exists pixel-wise misalignment between the estimated sketch and input sketch as they have different scales and viewpoints, we observe that the sketch consistency loss is robust to supervise the NeRF model to generate output that adheres to the sketch-specified geometry from arbitrary poses. 

\begin{figure*}[htb]
\begin{center}
\includegraphics[width=1.0\linewidth]{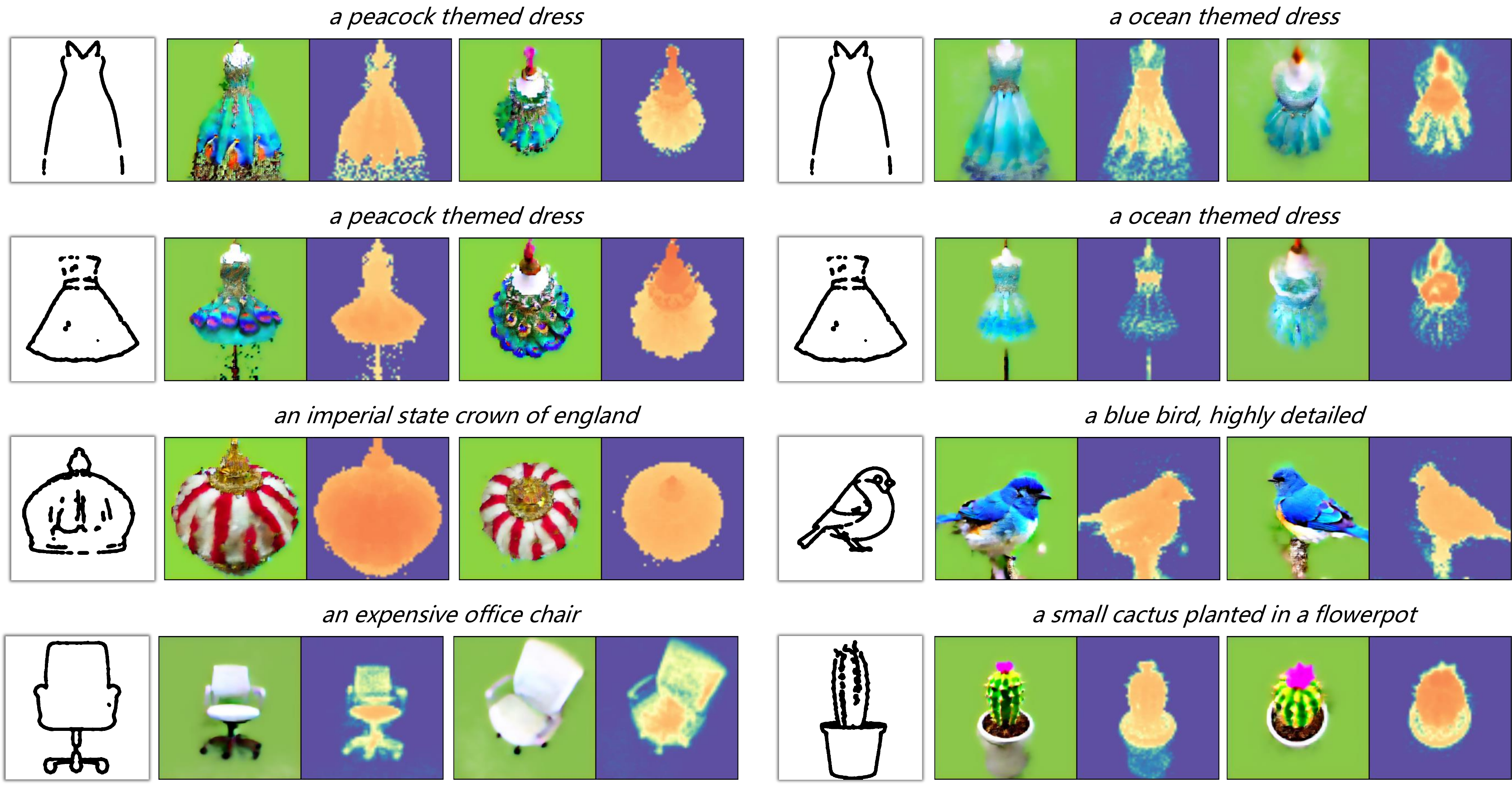}
\end{center}
\vspace{-0.1in}
\caption{Visualization of 3D assets generated by our Control3D. Given a hand-drawn sketch and a text prompt, our Control3D manages to generate plausible 3D contents which not only geometrically align with the input sketch but also faithfully match the input text prompts. Note that we visualize two different views of each 3D scene, and each view is coupled with the corresponding depth map.}
\label{fig:control3d_results}
\vspace{-0.2in}
\end{figure*}

\textbf{Overall Training.} Finally, the overall objective to train a NeRF for our controllable text-to-3D generation is given by:
\begin{equation}
\label{eq:total}
\mathcal{L}_{\text{total}} = \mathcal{L}_{C-SDS} + \mathcal{L}_{sketch}.
\end{equation}

\begin{figure*}[htbp]
\begin{center}
\includegraphics[width=1.0\linewidth]{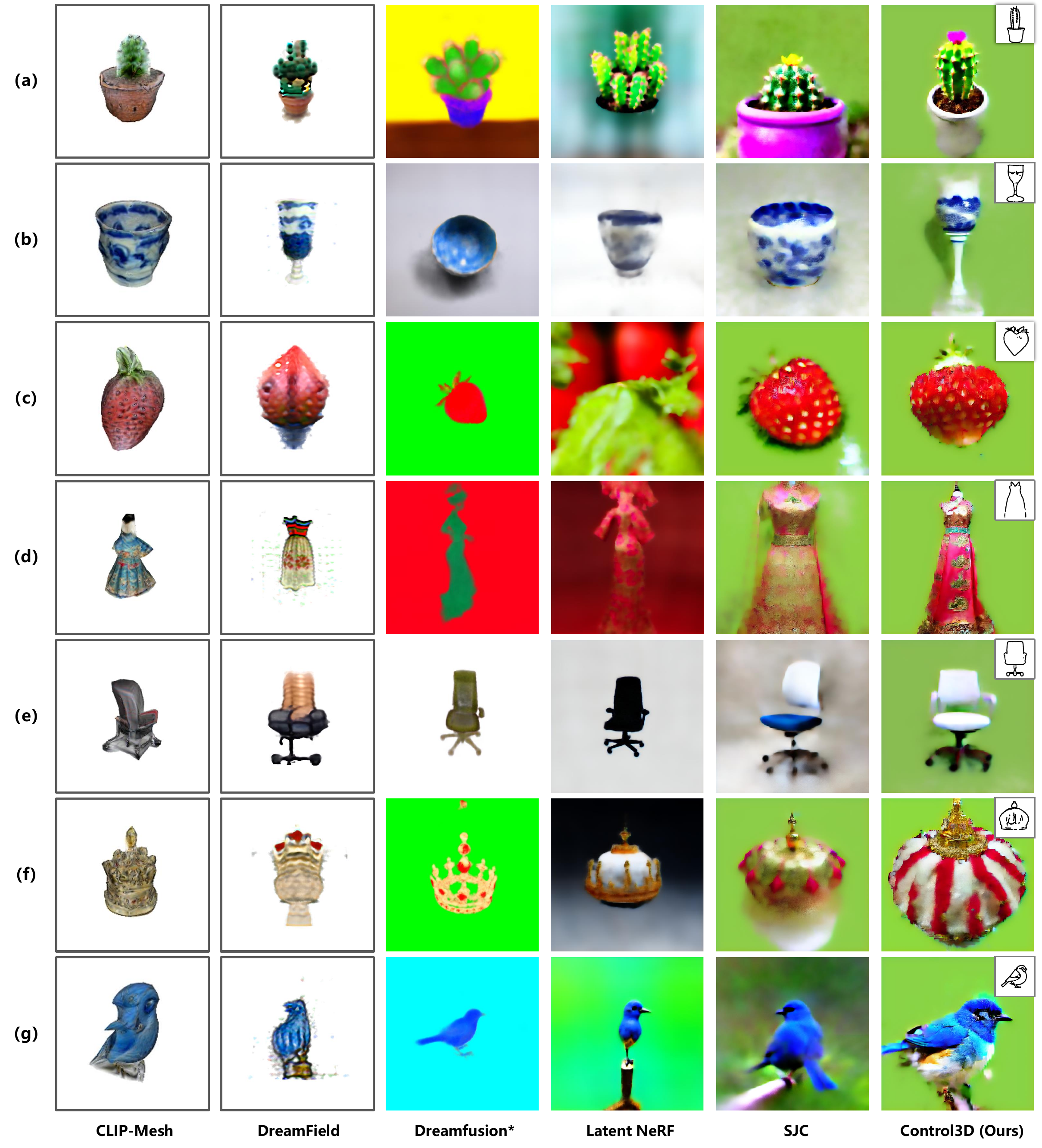}
\end{center}
\vspace{-0.15in}
\caption{Qualitative comparisons on text-to-3D generation. We compare our method with CLIP-Mesh \cite{mohammad2022clip},  DreamField \cite{jain2022zero},  DreamFusion* \cite{stable-dreamfusion}, Latent-NeRF \cite{metzer2022latent} and Score Jacobian Chaining (SJC) \cite{wang2022score}.  The prompts are (a) ``a small cactus planted in a flowerpot''; (b) ``a cup made of blue and white porcelain''; (c) ``a ripe strawberry''; (d) ``a Chinese style dress''; (e) ``an expensive office chair''; (f) ``an imperial state crown of england''; (g) ``a blue bird, highly detailed''. For each visualization of our Control3D (the last column), the sketch guidance used in the optimization process is also visualized in the upper right corner of the image. Our \textit{Control3D} shows better 3D results in terms of both geometry and texture in comparison to baselines.}
\label{fig:method_comparisons}
\end{figure*}

\section{Experiments}
\subsection{Implementation Details}
We implement the proposed \textit{Control3D} mainly based on the Score Jacobian Chaining (SJC) \cite{wang2022score} codebase. Following SJC, we use the voxel radiance field \cite{chen2022tensorf} to implement the underlying NeRF. SJC uses emptiness loss and center depth loss to regularize the NeRF learning. Our method also leverages these regularizations. For more details please refer to the original SJC \cite{wang2022score}. Following \cite{jain2021putting}, the image encoder used in Eq. \ref{eq:sketch} is the pre-trained CLIP ViT B/32 \cite{radford2021learning}. We resize the estimated sketch and input sketch to $224 \times 224$ resolution to match the input resolution of CLIP image encoder architecture. All experiments of \textit{Control3D} are conducted on a single NVIDIA V100 GPU. We train the model for 10,000 iterations and  the whole training process takes approximately an hour for each scene.

\subsection{Performance Comparison and Analysis}
\textbf{Visualization of Controllable Text-to-3D Generation via our Control3D.} Here we show the qualitative examples of our controllable text-to-3D generation with sketch guidance in Figure \ref{fig:control3d_results}. For each scene, we show several different views. In general, our Control3D manages to produce 3D scenes using simple hand-drawn sketches plus corresponding text prompts. We clearly observe that the synthetic 3D scenes faithfully respect both the semantic context present in the input text prompt and the geometric structure specified in the input sketch. Note that the input hand-drawn sketch is not required to be too strictly accurate or tight. Even when the contour curve of the input sketches only roughly describes the shapes of target 3D assets, our method can generate the corresponding results that basically align with the coarse geometry defined by the hand-drawn sketch (see the first row in Figure \ref{fig:control3d_results}. 

In addition, as shown in the first two rows of Figure \ref{fig:control3d_results}, given a fixed sketch, our method has the ability to generate different photo-realistic 3D scenes that conform to the corresponding text prompts. For instance, Control3D can generate ``peacock themed'' and ``ocean themed'' dresses that match the same input sketch by using different text prompts. Meanwhile, we also present another interesting case, by re-using the same text prompt and feeding different input sketches. As shown in the first two rows in Figure \ref{fig:control3d_results}, our method has the flexibility to demonstrate shape controls while preserving the same text-driven appearances. For example, Control3D is able to generate a ``full skirt'' and a ``midi skirt'' with the same appearance theme. The above observations demonstrate that our \textit{Control3D} may potentially enable many interesting 3D applications (such as recontextualization and reshaping), which would otherwise require tedious manual effort to tackle using traditional 3D modeling techniques.

\textbf{Qualitative Comparisons.}
To the best of our knowledge, our work is the first attempt to perform controllable text-to-3D generation with hand-drawn sketches. Hence, in the absence of an existing benchmark for comparison, we have to compare our method with existing text-to-3D generation methods which are solely conditioned on text prompts. Herein we compare our Control3D with five typical baselines. 1) CLIP-Mesh \cite{mohammad2022clip}, a zero-shot text-to-3D generation method using a pre-trained image-text model (i.e., CLIP \cite{radford2021learning}). 2) DreamField \cite{jain2022zero}, which combines neural radiance fields with CLIP to synthesis diverse 3D objects form text prompt. 3) DreamFusion*: As primary DreamFusion \cite{poole2022dreamfusion} leverages image diffusion priors from their private model Imagen \cite{saharia2022photorealistic}, we capitalize on the publicly available 2D diffusion model (Stable Diffusion) and reimplement DreamFusion based on \cite{stable-dreamfusion}, namely DreamFusion*. 4) Latent-NeRF \cite{metzer2022latent}, which learns a NeRF model on a latent feature space instead of in RGB pixel space, using a score distillation sampling loss in the latent space of Stable Diffusion. 5) Score Jacobian Chaining (SJC) \cite{wang2022score}, is another score distillation sampling baesd text-to-3D framework. It is worthy to note that the recent Magic3D \cite{lin2022magic3d} has shown high-quality text-to-3D generation results, it is excluded from comparisons since it relies on a private diffusion model eDiff-I \cite{balaji2022ediffi} that is unavailable to the research community.

\begin{figure*}[htb]
\begin{center}
\includegraphics[width=1.0\linewidth]{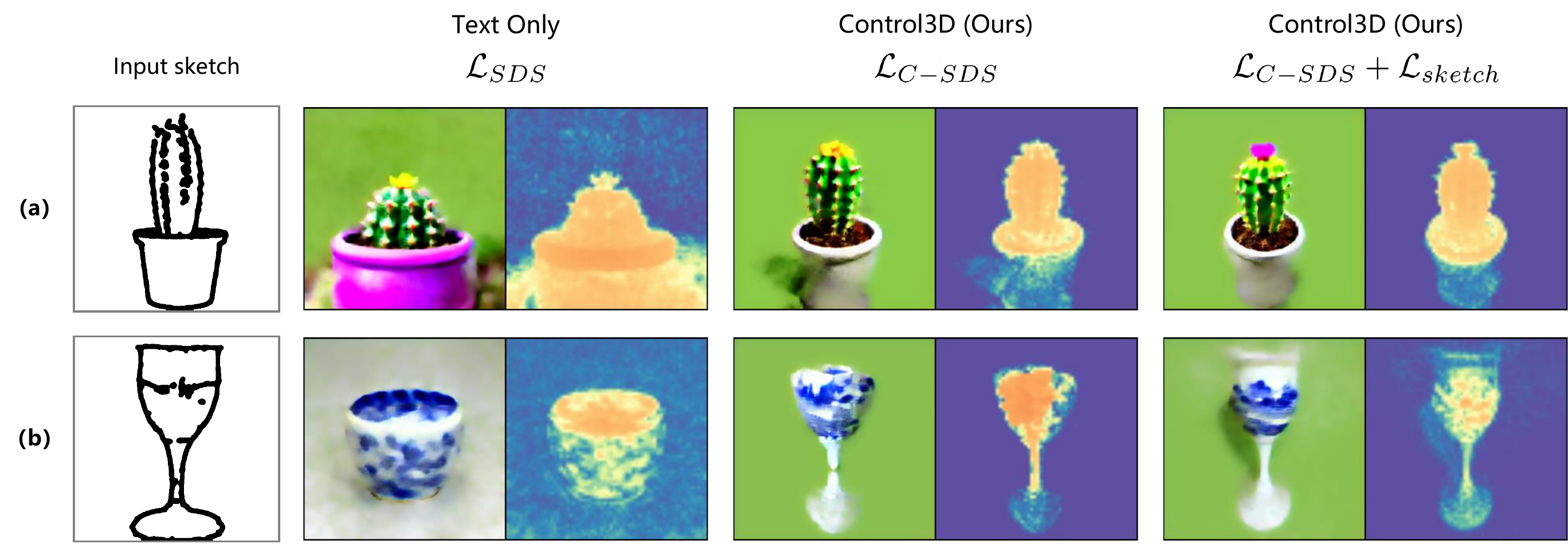}
\end{center}
\vspace{-0.15in}
\caption{Ablation studies of our Control3D. We demonstrate the effectiveness of each design in our Control3D for controllable text-to-3D generation. The prompts are (a) ``a small cactus planted in a flowerpot'' and (b) ``a cup made of blue and white porcelain''. Text-only ($\mathcal{L}_{SDS}$) is our baseline model SJC \cite{wang2022score} that creates 3D scenes only conditioned on text prompts. $\mathcal{L}_{C-SDS}$ is one variant of our Control3D that solely uses conditioned score sampling distillation loss. $\mathcal{L}_{C-SDS} + \mathcal{L}_{sketch}$ is the full version of Control3D that additionally leverages sketch consistency loss.}
\label{fig:ablation}
\vspace{-0.15in}
\end{figure*}

We depict the qualitative comparisons in Figure \ref{fig:method_comparisons}. As illustrated in this figure, CLIP-Mesh and DreamField show somewhat inferior capability of shape generation, making it difficult to generate plausible 3D shapes. Taking the fifth row (Figure \ref{fig:method_comparisons}(d)) as an example, when using the text prompt \emph{``an expensive office chair''}, CLIP-Mesh and DreamField generate chairs that are distorted and do not accurately match real-world chairs' structures. Although Dreamfusion* can generate reasonable 3D shapes, it encounters challenges in the generation of precise and realistic 3D textures, which consequently lead to unrealistic visual appearances. For instance, given the text prompt \emph{``an
imperial state crown of england''} and \emph{``blue bird, highly detailed''}, DreamFusion* can generate the accurate shapes of ``crown'' and ``bird'', but falls short in rendering fine texture details. While NeRFs operate in image space, DreamFusion* encodes rendered RGB images to a latent space in each and every training step for applying score distillation sampling with the publicly available Latent Diffusion Model (i.e., Stable Diffusion). Compare with the original DreamFusion which performs score distillation sampling in the standard RGB space by using their private RGB space diffusion model, this degraded guidance in latent space of DreamFusion* is somewhat insufficient and thus result in degenerated text-to-3D solutions. Instead, Latent-NeRF formulates the NeRF in the latent space, where the NeRF is optimized to render 2D feature maps in Stable Diffusion's latent space. These feature maps can easily be transformed back to RGB space through Stable Diffusion's image decoder. In this way, Latent-NeRF produces more textual details than DreamFusion*. However, Latent-NeRF's results still frequently suffer from blurry and diffuse issues.

In contrast, SJC and our \textit{Control3D} generate much better 3D structures than the aforementioned baselines. Furthermore, when compared to SJC, our \textit{Control3D} achieves higher 3D quality in terms of both geometry and texture. On the one hand, the hand-drawn sketch already depicts a well-drafted geometry and thus guides the NeRF model to generate plausible 3D shapes through our well-designed conditioned score sampling distillation and sketch consistency losses. On the other hand, with the sketch guidance, the text-conditioned probability density from the large-scale diffusion model has been narrowed down to a more compact region, which makes the underlying NeRF model easier to learn a  high-fidelity texture.
Accordingly, Our \textit{Control3D} manages to control text-to-3D generation with a human-drawn sketch, while all the baseline methods lack this ability. 

\textbf{User study.}
We additionally conducted a user study to quantitatively evaluate \textit{Control3D} against two diffusion based baseline models (i.e., Latent-NeRF and SJC) by comparing each pair. We invite 6 participants and show them two videos side by side in each test case. The videos are rendered from a canonical view by two different methods using the same text prompt. We then ask participants to choose the better one by jointly considering the following three aspects: (1) the alignment to the text prompt, (2) the fidelity of the visual appearance and (3) the accuracy of the geometry.
According to all participants’ feedback, we measure the user preference score of one method as the percentage of its generated results that are preferred. 
Table \ref{table:user_study} shows the results of the user study. In general, our \textit{Control3D} significantly outperforms the baseline methods with higher user preference rates.

\begin{table}[tb]
\caption{User Study. Users show a clear preference for Control3D over Latent-NeRF and SJC for text-to-3D generation.}
\centering
\vspace{-0.1in}
\begin{tabular}{lc}
\hline
Comparison & User Preference Score \\ \hline
Control3D \emph{vs.} Latent-NeRF &      81.2\%     \\
Control3D \emph{vs.} SJC  &      74.6\%     \\ \hline
\end{tabular}
\label{table:user_study}
\vspace{-0.2in}
\end{table}

\textbf{Ablation study.}
To enable controllable text-to-3D generation with a hand-drawn sketch, we design two loss terms: the conditioned score sampling distillation loss ($\mathcal{L}_{C-SDS}$) in Eq. \ref{eq:c-sds} and sketch consistency loss  ($\mathcal{L}_{sketch}$) in Eq. \ref{eq:sketch}. In this section, we investigate the effectiveness of each design. We depict the results of each ablated run in Figure \ref{fig:ablation}. Text-only is the base model SJC [38] that creates 3D scenes only adhering to the semantics of the input text prompt. Instead, when $\mathcal{L}_{C-SDS}$ is employed, the generated 3D scenes conform to both the input sketch and text prompt. This highlights the critical effectiveness of $\mathcal{L}_{C-SDS}$ for text-to-3D generation with sketch condition. However, when only $\mathcal{L}_{C-SDS}$ is applied, the generated 3D shapes don't precisely match the input sketch and may be distorted in local region. By utilizing an additional sketch consistency constraint $\mathcal{L}_{sketch}$, the shape mismatch and artifact issue is clearly alleviated. This demonstrates the advantage of our designed sketch consistency loss in Eq. \ref{eq:sketch}.

\section{Conclusion}
In this paper, we have proposed \textit{Control3D}, the first attempt to enhance user controllability in text-to-3d generation by incorporating hand-drawn sketch conditions. Specifically, a 2D conditioned diffusion model (ControlNet) is remoduled to optimize a Neural Radiance Field (NeRF), encouraging each view of the 3D scene to align with the given text prompt and hand-drawn sketch. Moreover, we propose a novel sketch consistency loss that explicitly encourages the geometric consistency between synthetic 3D scene and the given sketch. The extensive experiments demonstrate that the proposed method can generate accurate and faithful 3D scenes that closely align with the input text prompts and sketches. Our \textit{Control3D} provides a promising foundation for future research in controllable text-to-3D generation, which will lead to more creative and intuitive ways to generate 3D content. 


\begin{thebibliography}{47}
\balance

\ifx \showCODEN    \undefined \def \showCODEN     #1{\unskip}     \fi
\ifx \showDOI      \undefined \def \showDOI       #1{#1}\fi
\ifx \showISBNx    \undefined \def \showISBNx     #1{\unskip}     \fi
\ifx \showISBNxiii \undefined \def \showISBNxiii  #1{\unskip}     \fi
\ifx \showISSN     \undefined \def \showISSN      #1{\unskip}     \fi
\ifx \showLCCN     \undefined \def \showLCCN      #1{\unskip}     \fi
\ifx \shownote     \undefined \def \shownote      #1{#1}          \fi
\ifx \showarticletitle \undefined \def \showarticletitle #1{#1}   \fi
\ifx \showURL      \undefined \def \showURL       {\relax}        \fi
\providecommand\bibfield[2]{#2}
\providecommand\bibinfo[2]{#2}
\providecommand\natexlab[1]{#1}
\providecommand\showeprint[2][]{arXiv:#2}

\bibitem[Balaji et~al\mbox{.}(2022)]%
        {balaji2022ediffi}
\bibfield{author}{\bibinfo{person}{Yogesh Balaji}, \bibinfo{person}{Seungjun
  Nah}, \bibinfo{person}{Xun Huang}, \bibinfo{person}{Arash Vahdat},
  \bibinfo{person}{Jiaming Song}, \bibinfo{person}{Karsten Kreis},
  \bibinfo{person}{Miika Aittala}, \bibinfo{person}{Timo Aila},
  \bibinfo{person}{Samuli Laine}, \bibinfo{person}{Bryan Catanzaro},
  {et~al\mbox{.}}} \bibinfo{year}{2022}\natexlab{}.
\newblock \showarticletitle{ediffi: Text-to-image diffusion models with an
  ensemble of expert denoisers}.
\newblock \bibinfo{journal}{\emph{arXiv preprint arXiv:2211.01324}}
  (\bibinfo{year}{2022}).
\newblock


\bibitem[Chan et~al\mbox{.}(2022)]%
        {chan2022efficient}
\bibfield{author}{\bibinfo{person}{Eric~R Chan}, \bibinfo{person}{Connor~Z
  Lin}, \bibinfo{person}{Matthew~A Chan}, \bibinfo{person}{Koki Nagano},
  \bibinfo{person}{Boxiao Pan}, \bibinfo{person}{Shalini De~Mello},
  \bibinfo{person}{Orazio Gallo}, \bibinfo{person}{Leonidas~J Guibas},
  \bibinfo{person}{Jonathan Tremblay}, \bibinfo{person}{Sameh Khamis},
  {et~al\mbox{.}}} \bibinfo{year}{2022}\natexlab{}.
\newblock \showarticletitle{Efficient geometry-aware 3D generative adversarial
  networks}. In \bibinfo{booktitle}{\emph{CVPR}}.
\newblock


\bibitem[Chan et~al\mbox{.}(2021)]%
        {chan2021pi}
\bibfield{author}{\bibinfo{person}{Eric~R Chan}, \bibinfo{person}{Marco
  Monteiro}, \bibinfo{person}{Petr Kellnhofer}, \bibinfo{person}{Jiajun Wu},
  {and} \bibinfo{person}{Gordon Wetzstein}.} \bibinfo{year}{2021}\natexlab{}.
\newblock \showarticletitle{pi-gan: Periodic implicit generative adversarial
  networks for 3d-aware image synthesis}. In \bibinfo{booktitle}{\emph{CVPR}}.
\newblock


\bibitem[Chen et~al\mbox{.}(2022)]%
        {chen2022tensorf}
\bibfield{author}{\bibinfo{person}{Anpei Chen}, \bibinfo{person}{Zexiang Xu},
  \bibinfo{person}{Andreas Geiger}, \bibinfo{person}{Jingyi Yu}, {and}
  \bibinfo{person}{Hao Su}.} \bibinfo{year}{2022}\natexlab{}.
\newblock \showarticletitle{Tensorf: Tensorial radiance fields}. In
  \bibinfo{booktitle}{\emph{ECCV}}.
\newblock


\bibitem[Chen and Hays(2018)]%
        {chen2018sketchygan}
\bibfield{author}{\bibinfo{person}{Wengling Chen} {and} \bibinfo{person}{James
  Hays}.} \bibinfo{year}{2018}\natexlab{}.
\newblock \showarticletitle{Sketchygan: Towards diverse and realistic sketch to
  image synthesis}. In \bibinfo{booktitle}{\emph{CVPR}}.
\newblock


\bibitem[Chen et~al\mbox{.}(2019a)]%
        {chen2019animating}
\bibfield{author}{\bibinfo{person}{Yang Chen}, \bibinfo{person}{Yingwei Pan},
  \bibinfo{person}{Ting Yao}, \bibinfo{person}{Xinmei Tian}, {and}
  \bibinfo{person}{Tao Mei}.} \bibinfo{year}{2019}\natexlab{a}.
\newblock \showarticletitle{Animating Your Life: Real-Time Video-to-Animation
  Translation}. In \bibinfo{booktitle}{\emph{ACM MM Demo}}.
\newblock


\bibitem[Chen et~al\mbox{.}(2019b)]%
        {chen2019mocycle}
\bibfield{author}{\bibinfo{person}{Yang Chen}, \bibinfo{person}{Yingwei Pan},
  \bibinfo{person}{Ting Yao}, \bibinfo{person}{Xinmei Tian}, {and}
  \bibinfo{person}{Tao Mei}.} \bibinfo{year}{2019}\natexlab{b}.
\newblock \showarticletitle{Mocycle-gan: Unpaired video-to-video translation}.
  In \bibinfo{booktitle}{\emph{ACM MM}}.
\newblock


\bibitem[Deng et~al\mbox{.}(2022)]%
        {deng2022depth}
\bibfield{author}{\bibinfo{person}{Kangle Deng}, \bibinfo{person}{Andrew Liu},
  \bibinfo{person}{Jun-Yan Zhu}, {and} \bibinfo{person}{Deva Ramanan}.}
  \bibinfo{year}{2022}\natexlab{}.
\newblock \showarticletitle{Depth-supervised nerf: Fewer views and faster
  training for free}. In \bibinfo{booktitle}{\emph{CVPR}}.
\newblock


\bibitem[Gal et~al\mbox{.}(2023)]%
        {gal2022image}
\bibfield{author}{\bibinfo{person}{Rinon Gal}, \bibinfo{person}{Yuval Alaluf},
  \bibinfo{person}{Yuval Atzmon}, \bibinfo{person}{Or Patashnik},
  \bibinfo{person}{Amit~H Bermano}, \bibinfo{person}{Gal Chechik}, {and}
  \bibinfo{person}{Daniel Cohen-Or}.} \bibinfo{year}{2023}\natexlab{}.
\newblock \showarticletitle{An image is worth one word: Personalizing
  text-to-image generation using textual inversion}. In
  \bibinfo{booktitle}{\emph{ICLR}}.
\newblock


\bibitem[Gu et~al\mbox{.}(2022)]%
        {gu2021stylenerf}
\bibfield{author}{\bibinfo{person}{Jiatao Gu}, \bibinfo{person}{Lingjie Liu},
  \bibinfo{person}{Peng Wang}, {and} \bibinfo{person}{Christian Theobalt}.}
  \bibinfo{year}{2022}\natexlab{}.
\newblock \showarticletitle{Stylenerf: A style-based 3d-aware generator for
  high-resolution image synthesis}. In \bibinfo{booktitle}{\emph{ICLR}}.
\newblock


\bibitem[Guillard et~al\mbox{.}(2021)]%
        {guillard2021sketch2mesh}
\bibfield{author}{\bibinfo{person}{Benoit Guillard}, \bibinfo{person}{Edoardo
  Remelli}, \bibinfo{person}{Pierre Yvernay}, {and} \bibinfo{person}{Pascal
  Fua}.} \bibinfo{year}{2021}\natexlab{}.
\newblock \showarticletitle{Sketch2mesh: Reconstructing and editing 3d shapes
  from sketches}. In \bibinfo{booktitle}{\emph{ICCV}}.
\newblock


\bibitem[Ho et~al\mbox{.}(2020)]%
        {ho2020denoising}
\bibfield{author}{\bibinfo{person}{Jonathan Ho}, \bibinfo{person}{Ajay Jain},
  {and} \bibinfo{person}{Pieter Abbeel}.} \bibinfo{year}{2020}\natexlab{}.
\newblock \showarticletitle{Denoising diffusion probabilistic models}. In
  \bibinfo{booktitle}{\emph{NeurIPS}}.
\newblock


\bibitem[Ho and Salimans(2022)]%
        {ho2022classifier}
\bibfield{author}{\bibinfo{person}{Jonathan Ho} {and} \bibinfo{person}{Tim
  Salimans}.} \bibinfo{year}{2022}\natexlab{}.
\newblock \showarticletitle{Classifier-free diffusion guidance}. In
  \bibinfo{booktitle}{\emph{NeurIPS Workshop}}.
\newblock


\bibitem[Jain et~al\mbox{.}(2022)]%
        {jain2022zero}
\bibfield{author}{\bibinfo{person}{Ajay Jain}, \bibinfo{person}{Ben
  Mildenhall}, \bibinfo{person}{Jonathan~T Barron}, \bibinfo{person}{Pieter
  Abbeel}, {and} \bibinfo{person}{Ben Poole}.} \bibinfo{year}{2022}\natexlab{}.
\newblock \showarticletitle{Zero-shot text-guided object generation with dream
  fields}. In \bibinfo{booktitle}{\emph{CVPR}}.
\newblock


\bibitem[Jain et~al\mbox{.}(2021)]%
        {jain2021putting}
\bibfield{author}{\bibinfo{person}{Ajay Jain}, \bibinfo{person}{Matthew
  Tancik}, {and} \bibinfo{person}{Pieter Abbeel}.}
  \bibinfo{year}{2021}\natexlab{}.
\newblock \showarticletitle{Putting nerf on a diet: Semantically consistent
  few-shot view synthesis}. In \bibinfo{booktitle}{\emph{ICCV}}.
\newblock


\bibitem[Kim et~al\mbox{.}(2022)]%
        {kim2022infonerf}
\bibfield{author}{\bibinfo{person}{Mijeong Kim}, \bibinfo{person}{Seonguk Seo},
  {and} \bibinfo{person}{Bohyung Han}.} \bibinfo{year}{2022}\natexlab{}.
\newblock \showarticletitle{Infonerf: Ray entropy minimization for few-shot
  neural volume rendering}. In \bibinfo{booktitle}{\emph{CVPR}}.
\newblock


\bibitem[Li et~al\mbox{.}(2019)]%
        {li2019photo}
\bibfield{author}{\bibinfo{person}{Mengtian Li}, \bibinfo{person}{Zhe Lin},
  \bibinfo{person}{Radomir Mech}, \bibinfo{person}{Ersin Yumer}, {and}
  \bibinfo{person}{Deva Ramanan}.} \bibinfo{year}{2019}\natexlab{}.
\newblock \showarticletitle{Photo-sketching: Inferring contour drawings from
  images}. In \bibinfo{booktitle}{\emph{WACV}}.
\newblock


\bibitem[Li et~al\mbox{.}(2022)]%
        {li2022contextual}
\bibfield{author}{\bibinfo{person}{Yehao Li}, \bibinfo{person}{Ting Yao},
  \bibinfo{person}{Yingwei Pan}, {and} \bibinfo{person}{Tao Mei}.}
  \bibinfo{year}{2022}\natexlab{}.
\newblock \showarticletitle{Contextual transformer networks for visual
  recognition}.
\newblock \bibinfo{journal}{\emph{IEEE TPAMI}} (\bibinfo{year}{2022}).
\newblock


\bibitem[Liao et~al\mbox{.}(2020)]%
        {liao2020towards}
\bibfield{author}{\bibinfo{person}{Yiyi Liao}, \bibinfo{person}{Katja Schwarz},
  \bibinfo{person}{Lars Mescheder}, {and} \bibinfo{person}{Andreas Geiger}.}
  \bibinfo{year}{2020}\natexlab{}.
\newblock \showarticletitle{Towards unsupervised learning of generative models
  for 3d controllable image synthesis}. In \bibinfo{booktitle}{\emph{CVPR}}.
\newblock


\bibitem[Lin et~al\mbox{.}(2023)]%
        {lin2022magic3d}
\bibfield{author}{\bibinfo{person}{Chen-Hsuan Lin}, \bibinfo{person}{Jun Gao},
  \bibinfo{person}{Luming Tang}, \bibinfo{person}{Towaki Takikawa},
  \bibinfo{person}{Xiaohui Zeng}, \bibinfo{person}{Xun Huang},
  \bibinfo{person}{Karsten Kreis}, \bibinfo{person}{Sanja Fidler},
  \bibinfo{person}{Ming-Yu Liu}, {and} \bibinfo{person}{Tsung-Yi Lin}.}
  \bibinfo{year}{2023}\natexlab{}.
\newblock \showarticletitle{Magic3D: High-Resolution Text-to-3D Content
  Creation}. In \bibinfo{booktitle}{\emph{CVPR}}.
\newblock


\bibitem[Lu et~al\mbox{.}(2018)]%
        {lu2018image}
\bibfield{author}{\bibinfo{person}{Yongyi Lu}, \bibinfo{person}{Shangzhe Wu},
  \bibinfo{person}{Yu-Wing Tai}, {and} \bibinfo{person}{Chi-Keung Tang}.}
  \bibinfo{year}{2018}\natexlab{}.
\newblock \showarticletitle{Image generation from sketch constraint using
  contextual gan}. In \bibinfo{booktitle}{\emph{ECCV}}.
\newblock


\bibitem[Max(1995)]%
        {max1995optical}
\bibfield{author}{\bibinfo{person}{Nelson Max}.}
  \bibinfo{year}{1995}\natexlab{}.
\newblock \showarticletitle{Optical models for direct volume rendering}.
\newblock \bibinfo{journal}{\emph{IEEE TVCG}} (\bibinfo{year}{1995}).
\newblock


\bibitem[Metzer et~al\mbox{.}(2023)]%
        {metzer2022latent}
\bibfield{author}{\bibinfo{person}{Gal Metzer}, \bibinfo{person}{Elad
  Richardson}, \bibinfo{person}{Or Patashnik}, \bibinfo{person}{Raja Giryes},
  {and} \bibinfo{person}{Daniel Cohen-Or}.} \bibinfo{year}{2023}\natexlab{}.
\newblock \showarticletitle{Latent-NeRF for Shape-Guided Generation of 3D
  Shapes and Textures}. In \bibinfo{booktitle}{\emph{CVPR}}.
\newblock


\bibitem[Mildenhall et~al\mbox{.}(2020)]%
        {mildenhall2020nerf}
\bibfield{author}{\bibinfo{person}{Ben Mildenhall}, \bibinfo{person}{Pratul~P
  Srinivasan}, \bibinfo{person}{Matthew Tancik}, \bibinfo{person}{Jonathan~T
  Barron}, \bibinfo{person}{Ravi Ramamoorthi}, {and} \bibinfo{person}{Ren Ng}.}
  \bibinfo{year}{2020}\natexlab{}.
\newblock \showarticletitle{{NeRF: Representing Scenes as Neural Radiance
  Fields for View Synthesis}}. In \bibinfo{booktitle}{\emph{ECCV}}.
\newblock


\bibitem[Mohammad~Khalid et~al\mbox{.}(2022)]%
        {mohammad2022clip}
\bibfield{author}{\bibinfo{person}{Nasir Mohammad~Khalid},
  \bibinfo{person}{Tianhao Xie}, \bibinfo{person}{Eugene Belilovsky}, {and}
  \bibinfo{person}{Tiberiu Popa}.} \bibinfo{year}{2022}\natexlab{}.
\newblock \showarticletitle{CLIP-Mesh: Generating textured meshes from text
  using pretrained image-text models}. In \bibinfo{booktitle}{\emph{SIGGRAPH
  Asia}}.
\newblock


\bibitem[Nguyen-Phuoc et~al\mbox{.}(2019)]%
        {nguyen2019hologan}
\bibfield{author}{\bibinfo{person}{Thu Nguyen-Phuoc}, \bibinfo{person}{Chuan
  Li}, \bibinfo{person}{Lucas Theis}, \bibinfo{person}{Christian Richardt},
  {and} \bibinfo{person}{Yong-Liang Yang}.} \bibinfo{year}{2019}\natexlab{}.
\newblock \showarticletitle{Hologan: Unsupervised learning of 3d
  representations from natural images}. In \bibinfo{booktitle}{\emph{ICCV}}.
\newblock


\bibitem[Nichol and Dhariwal(2021)]%
        {nichol2021improved}
\bibfield{author}{\bibinfo{person}{Alexander~Quinn Nichol} {and}
  \bibinfo{person}{Prafulla Dhariwal}.} \bibinfo{year}{2021}\natexlab{}.
\newblock \showarticletitle{Improved denoising diffusion probabilistic models}.
  In \bibinfo{booktitle}{\emph{ICLR}}.
\newblock


\bibitem[Niemeyer et~al\mbox{.}(2022)]%
        {niemeyer2022regnerf}
\bibfield{author}{\bibinfo{person}{Michael Niemeyer},
  \bibinfo{person}{Jonathan~T Barron}, \bibinfo{person}{Ben Mildenhall},
  \bibinfo{person}{Mehdi~SM Sajjadi}, \bibinfo{person}{Andreas Geiger}, {and}
  \bibinfo{person}{Noha Radwan}.} \bibinfo{year}{2022}\natexlab{}.
\newblock \showarticletitle{Regnerf: Regularizing neural radiance fields for
  view synthesis from sparse inputs}. In \bibinfo{booktitle}{\emph{CVPR}}.
\newblock


\bibitem[Niemeyer and Geiger(2021)]%
        {niemeyer2021giraffe}
\bibfield{author}{\bibinfo{person}{Michael Niemeyer} {and}
  \bibinfo{person}{Andreas Geiger}.} \bibinfo{year}{2021}\natexlab{}.
\newblock \showarticletitle{Giraffe: Representing scenes as compositional
  generative neural feature fields}. In \bibinfo{booktitle}{\emph{CVPR}}.
\newblock


\bibitem[Pan et~al\mbox{.}(2017)]%
        {pan2017create}
\bibfield{author}{\bibinfo{person}{Yingwei Pan}, \bibinfo{person}{Zhaofan Qiu},
  \bibinfo{person}{Ting Yao}, \bibinfo{person}{Houqiang Li}, {and}
  \bibinfo{person}{Tao Mei}.} \bibinfo{year}{2017}\natexlab{}.
\newblock \showarticletitle{To create what you tell: Generating videos from
  captions}. In \bibinfo{booktitle}{\emph{ACM MM}}.
\newblock


\bibitem[Poole et~al\mbox{.}(2023)]%
        {poole2022dreamfusion}
\bibfield{author}{\bibinfo{person}{Ben Poole}, \bibinfo{person}{Ajay Jain},
  \bibinfo{person}{Jonathan~T Barron}, {and} \bibinfo{person}{Ben Mildenhall}.}
  \bibinfo{year}{2023}\natexlab{}.
\newblock \showarticletitle{Dreamfusion: Text-to-3d using 2d diffusion}. In
  \bibinfo{booktitle}{\emph{ICLR}}.
\newblock


\bibitem[Radford et~al\mbox{.}(2021)]%
        {radford2021learning}
\bibfield{author}{\bibinfo{person}{Alec Radford}, \bibinfo{person}{Jong~Wook
  Kim}, \bibinfo{person}{Chris Hallacy}, \bibinfo{person}{Aditya Ramesh},
  \bibinfo{person}{Gabriel Goh}, \bibinfo{person}{Sandhini Agarwal},
  \bibinfo{person}{Girish Sastry}, \bibinfo{person}{Amanda Askell},
  \bibinfo{person}{Pamela Mishkin}, \bibinfo{person}{Jack Clark},
  {et~al\mbox{.}}} \bibinfo{year}{2021}\natexlab{}.
\newblock \showarticletitle{Learning transferable visual models from natural
  language supervision}. In \bibinfo{booktitle}{\emph{ICML}}.
\newblock


\bibitem[Ramesh et~al\mbox{.}(2022)]%
        {ramesh2022hierarchical}
\bibfield{author}{\bibinfo{person}{Aditya Ramesh}, \bibinfo{person}{Prafulla
  Dhariwal}, \bibinfo{person}{Alex Nichol}, \bibinfo{person}{Casey Chu}, {and}
  \bibinfo{person}{Mark Chen}.} \bibinfo{year}{2022}\natexlab{}.
\newblock \showarticletitle{Hierarchical text-conditional image generation with
  clip latents}.
\newblock \bibinfo{journal}{\emph{arXiv preprint arXiv:2204.06125}}
  (\bibinfo{year}{2022}).
\newblock


\bibitem[Rombach et~al\mbox{.}(2022)]%
        {rombach2022high}
\bibfield{author}{\bibinfo{person}{Robin Rombach}, \bibinfo{person}{Andreas
  Blattmann}, \bibinfo{person}{Dominik Lorenz}, \bibinfo{person}{Patrick
  Esser}, {and} \bibinfo{person}{Bj{\"o}rn Ommer}.}
  \bibinfo{year}{2022}\natexlab{}.
\newblock \showarticletitle{High-resolution image synthesis with latent
  diffusion models}. In \bibinfo{booktitle}{\emph{CVPR}}.
\newblock


\bibitem[Ruiz et~al\mbox{.}(2023)]%
        {ruiz2022dreambooth}
\bibfield{author}{\bibinfo{person}{Nataniel Ruiz}, \bibinfo{person}{Yuanzhen
  Li}, \bibinfo{person}{Varun Jampani}, \bibinfo{person}{Yael Pritch},
  \bibinfo{person}{Michael Rubinstein}, {and} \bibinfo{person}{Kfir Aberman}.}
  \bibinfo{year}{2023}\natexlab{}.
\newblock \showarticletitle{Dreambooth: Fine tuning text-to-image diffusion
  models for subject-driven generation}. In \bibinfo{booktitle}{\emph{CVPR}}.
\newblock


\bibitem[Saharia et~al\mbox{.}(2022)]%
        {saharia2022photorealistic}
\bibfield{author}{\bibinfo{person}{Chitwan Saharia}, \bibinfo{person}{William
  Chan}, \bibinfo{person}{Saurabh Saxena}, \bibinfo{person}{Lala Li},
  \bibinfo{person}{Jay Whang}, \bibinfo{person}{Emily~L Denton},
  \bibinfo{person}{Kamyar Ghasemipour}, \bibinfo{person}{Raphael
  Gontijo~Lopes}, \bibinfo{person}{Burcu Karagol~Ayan}, \bibinfo{person}{Tim
  Salimans}, {et~al\mbox{.}}} \bibinfo{year}{2022}\natexlab{}.
\newblock \showarticletitle{Photorealistic text-to-image diffusion models with
  deep language understanding}. In \bibinfo{booktitle}{\emph{NeurIPS}}.
\newblock


\bibitem[Sain et~al\mbox{.}(2023)]%
        {sain2023clip}
\bibfield{author}{\bibinfo{person}{Aneeshan Sain}, \bibinfo{person}{Ayan~Kumar
  Bhunia}, \bibinfo{person}{Pinaki~Nath Chowdhury}, \bibinfo{person}{Subhadeep
  Koley}, \bibinfo{person}{Tao Xiang}, {and} \bibinfo{person}{Yi-Zhe Song}.}
  \bibinfo{year}{2023}\natexlab{}.
\newblock \showarticletitle{CLIP for All Things Zero-Shot Sketch-Based Image
  Retrieval, Fine-Grained or Not}. In \bibinfo{booktitle}{\emph{CVPR}}.
\newblock


\bibitem[Sohl-Dickstein et~al\mbox{.}(2015)]%
        {sohl2015deep}
\bibfield{author}{\bibinfo{person}{Jascha Sohl-Dickstein},
  \bibinfo{person}{Eric Weiss}, \bibinfo{person}{Niru Maheswaranathan}, {and}
  \bibinfo{person}{Surya Ganguli}.} \bibinfo{year}{2015}\natexlab{}.
\newblock \showarticletitle{Deep unsupervised learning using nonequilibrium
  thermodynamics}. In \bibinfo{booktitle}{\emph{ICML}}.
\newblock


\bibitem[Sutherland(1964)]%
        {sutherland1964sketch}
\bibfield{author}{\bibinfo{person}{Ivan~E Sutherland}.}
  \bibinfo{year}{1964}\natexlab{}.
\newblock \showarticletitle{Sketch pad a man-machine graphical communication
  system}. In \bibinfo{booktitle}{\emph{Proceedings of the SHARE design
  automation workshop}}.
\newblock


\bibitem[Tang(2022)]%
        {stable-dreamfusion}
\bibfield{author}{\bibinfo{person}{Jiaxiang Tang}.}
  \bibinfo{year}{2022}\natexlab{}.
\newblock \bibinfo{title}{Stable-dreamfusion: Text-to-3D with
  Stable-diffusion}.
\newblock
\newblock
\newblock
\shownote{https://github.com/ashawkey/stable-dreamfusion}.


\bibitem[Voynov et~al\mbox{.}(2022)]%
        {voynov2022sketch}
\bibfield{author}{\bibinfo{person}{Andrey Voynov}, \bibinfo{person}{Kfir
  Aberman}, {and} \bibinfo{person}{Daniel Cohen-Or}.}
  \bibinfo{year}{2022}\natexlab{}.
\newblock \showarticletitle{Sketch-Guided Text-to-Image Diffusion Models}.
\newblock \bibinfo{journal}{\emph{arXiv preprint arXiv:2211.13752}}
  (\bibinfo{year}{2022}).
\newblock


\bibitem[Wang et~al\mbox{.}(2023)]%
        {wang2022score}
\bibfield{author}{\bibinfo{person}{Haochen Wang}, \bibinfo{person}{Xiaodan Du},
  \bibinfo{person}{Jiahao Li}, \bibinfo{person}{Raymond~A Yeh}, {and}
  \bibinfo{person}{Greg Shakhnarovich}.} \bibinfo{year}{2023}\natexlab{}.
\newblock \showarticletitle{Score Jacobian Chaining: Lifting Pretrained 2D
  Diffusion Models for 3D Generation}. In \bibinfo{booktitle}{\emph{CVPR}}.
\newblock


\bibitem[Xu et~al\mbox{.}(2022)]%
        {xu2022sinnerf}
\bibfield{author}{\bibinfo{person}{Dejia Xu}, \bibinfo{person}{Yifan Jiang},
  \bibinfo{person}{Peihao Wang}, \bibinfo{person}{Zhiwen Fan},
  \bibinfo{person}{Humphrey Shi}, {and} \bibinfo{person}{Zhangyang Wang}.}
  \bibinfo{year}{2022}\natexlab{}.
\newblock \showarticletitle{Sinnerf: Training neural radiance fields on complex
  scenes from a single image}. In \bibinfo{booktitle}{\emph{ECCV}}.
\newblock


\bibitem[Yao et~al\mbox{.}(2023)]%
        {yao2023dual}
\bibfield{author}{\bibinfo{person}{Ting Yao}, \bibinfo{person}{Yehao Li},
  \bibinfo{person}{Yingwei Pan}, \bibinfo{person}{Yu Wang},
  \bibinfo{person}{Xiao-Ping Zhang}, {and} \bibinfo{person}{Tao Mei}.}
  \bibinfo{year}{2023}\natexlab{}.
\newblock \showarticletitle{Dual vision transformer}.
\newblock \bibinfo{journal}{\emph{IEEE TPAMI}} (\bibinfo{year}{2023}).
\newblock


\bibitem[Yao et~al\mbox{.}(2022)]%
        {yao2022wave}
\bibfield{author}{\bibinfo{person}{Ting Yao}, \bibinfo{person}{Yingwei Pan},
  \bibinfo{person}{Yehao Li}, \bibinfo{person}{Chong-Wah Ngo}, {and}
  \bibinfo{person}{Tao Mei}.} \bibinfo{year}{2022}\natexlab{}.
\newblock \showarticletitle{Wave-vit: Unifying wavelet and transformers for
  visual representation learning}. In \bibinfo{booktitle}{\emph{ECCV}}.
\newblock


\bibitem[Zhang and Agrawala(2023)]%
        {zhang2023adding}
\bibfield{author}{\bibinfo{person}{Lvmin Zhang} {and} \bibinfo{person}{Maneesh
  Agrawala}.} \bibinfo{year}{2023}\natexlab{}.
\newblock \showarticletitle{Adding conditional control to text-to-image
  diffusion models}.
\newblock \bibinfo{journal}{\emph{arXiv preprint arXiv:2302.05543}}
  (\bibinfo{year}{2023}).
\newblock


\bibitem[Zhang et~al\mbox{.}(2023)]%
        {zhang2023}
\bibfield{author}{\bibinfo{person}{Yiheng Zhang}, \bibinfo{person}{Zhaofan
  Qiu}, \bibinfo{person}{Yingwei Pan}, \bibinfo{person}{Ting Yao}, {and}
  \bibinfo{person}{Tao Mei}.} \bibinfo{year}{2023}\natexlab{}.
\newblock \showarticletitle{Learning Neural Implicit Surfaces with Object-Aware
  Radiance Fields}. In \bibinfo{booktitle}{\emph{ICCV}}.
\newblock


\end{thebibliography}

\end{document}